\documentclass[10pt,twocolumn,letterpaper]{article}

\usepackage{cvpr}              

\usepackage{graphicx}
\usepackage{amsmath}
\usepackage{amssymb}
\usepackage{booktabs}
\usepackage{bbding}

\usepackage{algorithm} 
\usepackage{algorithmic}  
\usepackage[algo2e,ruled,linesnumbered]{algorithm2e}
\SetKwInOut{Parameter}{parameter}

\SetCommentSty{mycommfont}

\usepackage{xcolor}

\newcommand*\samethanks[1][\value{footnote}]{\footnotemark[#1]}

\def\cvprPaperID{3666} 
\def\confName{CVPR}
\def\confYear{2023}

\usepackage[colorlinks,linkcolor=red]{hyperref}
\usepackage[capitalize]{cleveref}

\begin{document}

\crefname{section}{Sec.}{Secs.}
\Crefname{section}{Section}{Sections}
\Crefname{table}{Table}{Tables}
\crefname{table}{Tab.}{Tabs.}

\title{OTAvatar$\colon$One-shot Talking Face Avatar with Controllable Tri-plane Rendering}

\author{{Zhiyuan Ma$^{1,2}$\thanks{Equal contribution.} \qquad Xiangyu Zhu$^{3}$\samethanks \qquad Guojun Qi$^4$ \qquad Zhen Lei$^{1,2,3}$\thanks{Corresponding author.} \qquad Lei Zhang$^{1}$ } \\
$^1$The Hong Kong Polytechnic University \\ 
$^2$Center for Artificial Intelligence and Robotics, HKISI CAS \\
$^3$State Key Laboratory of Multimodal Artificial Intelligence Systems, CASIA \\
$^4$OPPO Research \\
 \tt\small \ zm2354.ma@connect.polyu.hk, xiangyu.zhu@nlpr.ia.ac.cn \\ \tt\small guojunq@gmail.com, zlei@nlpr.ia.ac.cn, cslzhang@comp.polyu.edu.hk
}
\maketitle

\begin{abstract}

Controllability, generalizability and efficiency are the major objectives of constructing face avatars represented by neural implicit field. However, existing methods have not managed to accommodate the three requirements simultaneously. They either focus on static portraits, restricting the representation ability to a specific subject, or suffer from substantial computational cost, limiting their flexibility. In this paper, we propose One-shot Talking face Avatar (OTAvatar), which constructs face avatars by a generalized controllable tri-plane rendering solution so that each personalized avatar can be constructed from only one portrait as the reference. Specifically, OTAvatar first inverts a portrait image to a motion-free identity code. Second, the identity code and a motion code are utilized to modulate an efficient CNN to generate a tri-plane formulated volume, which encodes the subject in the desired motion. Finally, volume rendering is employed to generate an image in any view. The core of our solution is a novel decoupling-by-inverting strategy that disentangles identity and motion in the latent code via optimization-based inversion. Benefiting from the efficient tri-plane representation, we achieve controllable rendering of generalized face avatar at $35$ FPS on A100. Experiments show promising performance of cross-identity reenactment on subjects out of the training set and better 3D consistency. \textcolor{black}{The code is available at} \href{https://github.com/theEricMa/OTAvatar}{https://github.com/theEricMa/OTAvatar}.

\end{abstract}

\begin{figure}[ht]
  \centering
   \includegraphics[width=1\linewidth]{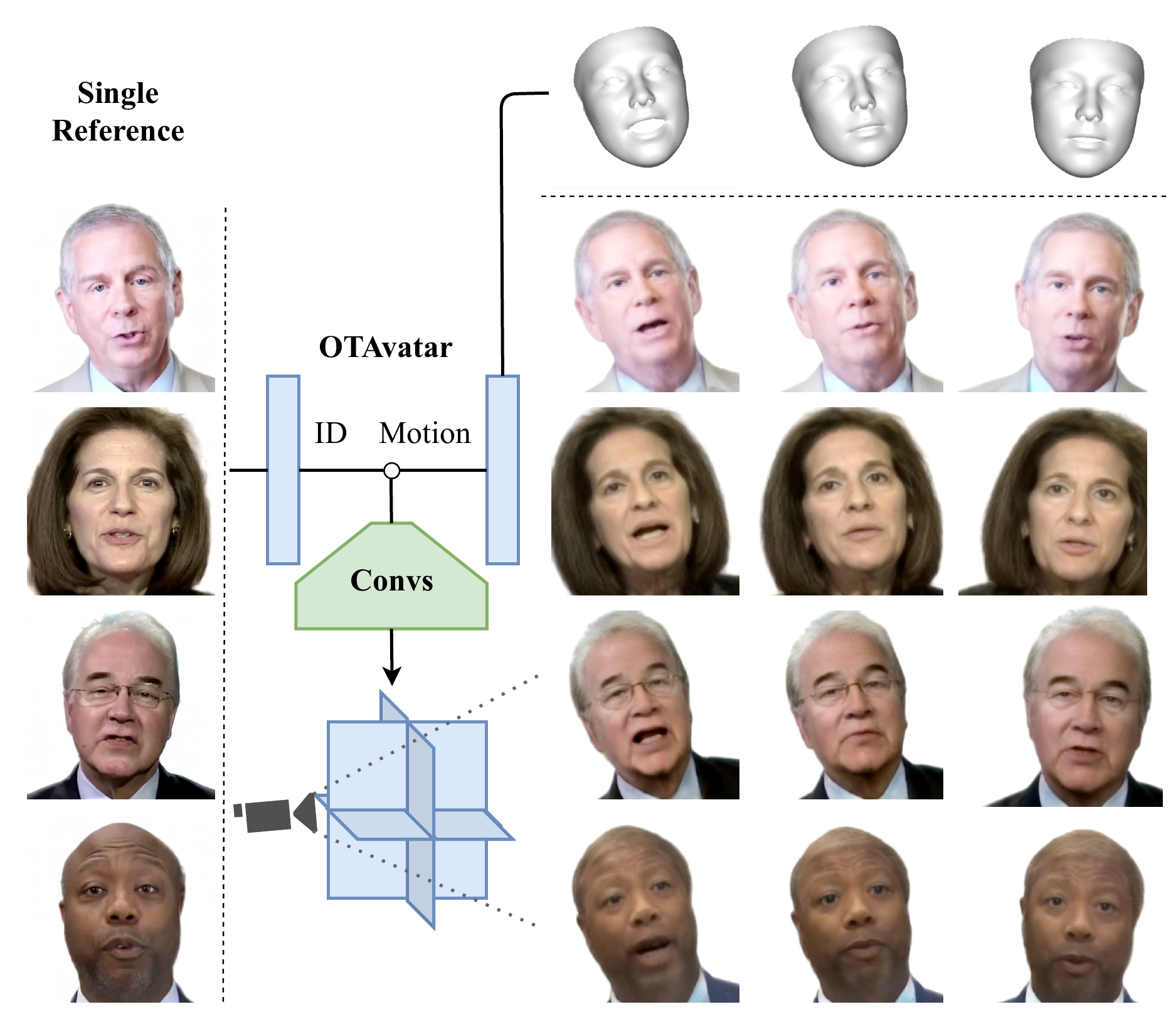}

   \caption{\textbf{OTAvatar animation results}. The source subjects in HDTF\cite{zhang2021flow} dataset are animated by OTAvatar using a single portrait as the reference. We use the pose and expression coefficients of 3DMM to represent motion and drive the avatar. Note that these subjects are \textbf{not included} in the training data of OTAvatar.}
   \label{fig:animation result}
\end{figure}

\section{Introduction}
Neural rendering has achieved remarkable progress and promising results in 3D reconstruction. Thanks to the differentiability in neuron computation, the neural rendering methods bypass the expensive cost of high-fidelity digital 3D modeling, thus attracting the attention of many researchers from academia and industry. In this work, we aim to generate a talking face avatar via neural rendering techniques, which is controllable by driving signals like video and audio segments. Such animation is expected to inherit the identity from reference portraits, while the expression and pose can keep in sync with the driving signals.

The early works about talking face focus on expression animation consistent with driving signals in constrained frontal face images~\cite{chung2016lip, prajwal2020lip, chen2019hierarchical}. It is then extended to in-the-wild scenes with pose variations
~\cite{ji2021audio,wang2018vid2vid, zhou2021pose, zhou2018talking}. 
Many previous works are 2D methods, where the image warping adopted to stimulate the motion in 3D space is learned in 2D space~\cite{ren2021pirenderer, siarohin2019first, siarohin2021motion, wang2022latent, yin2022styleheat}. These methods tend to collapse under large pose variation. In contrast, there are 3D methods that can address the pose problem, since the head pose can be treated as a novel view. Both explicit \cite{fernando2004gpu, levoy1988display} and implicit 3D representations\cite{mildenhall2021nerf} are introduced to face rendering\cite{Lombardi:2019, raj2021pva, Mihajlovic:KeypointNeRF:ECCV2022, park2021nerfies, guo2021ad, zheng2022imavatar, hong2022headnerf}. However, these methods either overfit to a single identity or fail to produce high-quality animation for different identities. 

In this work, we propose a one-shot talking face avatar (OTAvatar), which can generate mimic expressions with good 3D consistency and be generalized to different identities with only one portrait reference. Some avatar animation examples are shown in Fig.~\ref{fig:animation result}. Given a single reference image, OTAvatar can drive the subject with motion signals to generate corresponding face images. We realize it under the framework of volume rendering. Current methods usually render static subjects~\cite{Mihajlovic:KeypointNeRF:ECCV2022, raj2021pva}. Although there are works~\cite{guo2021ad,zheng2022imavatar,park2021nerfies, park2021hypernerf,Gafni_2021_CVPR} proposed to implement dynamic rendering for face avatars, they need one model per subject. Therefore, the generalization is poor. HeadNeRF\cite{hong2022headnerf} is a similar work to us. However, its reconstruction results are unnatural in the talking face case. 

Our method is built on a 3D generative model pre-trained on a large-scale face database to guarantee identity generalization ability. Besides, we employ a motion controller to decouple the motion and identity in latent space when performing the optimization-based GAN inversion, to make the motion controllable and transferable to different identities. The network architecture is compact to ensure inference efficiency. In the utilization of our OTAvatar, given a single reference image, we fix the motion code predicted by the controller and only optimize the identity code so that a new avatar of the reference image can be constructed. Such a disentanglement enables the rendering of any desired motion by simply alternating the motion-related latent code to be that of the specific motion representation.

 \vspace{+3mm}
The major contributions of this work can be summarized as follows.
\begin{itemize}
    \item We make the first attempt for one-shot 3D face reconstruction and motion-controllable rendering by taming a pre-trained 3D generative model for motion control. 

    \item We propose to decouple motion-related and motion-free latent code in inversion optimization by prompting the motion fraction of latent code ahead of the optimization using a decoupling-by-inverting strategy. 

    \item Our method can photo-realistically render any identity with the desired expression and pose at 35FPS. The experiment shows promising results of natural motion and 3D consistency on both 2D and 3D datasets. 
    
\end{itemize}

\begin{figure*}[ht]
  \centering
   \includegraphics[width=1\linewidth]{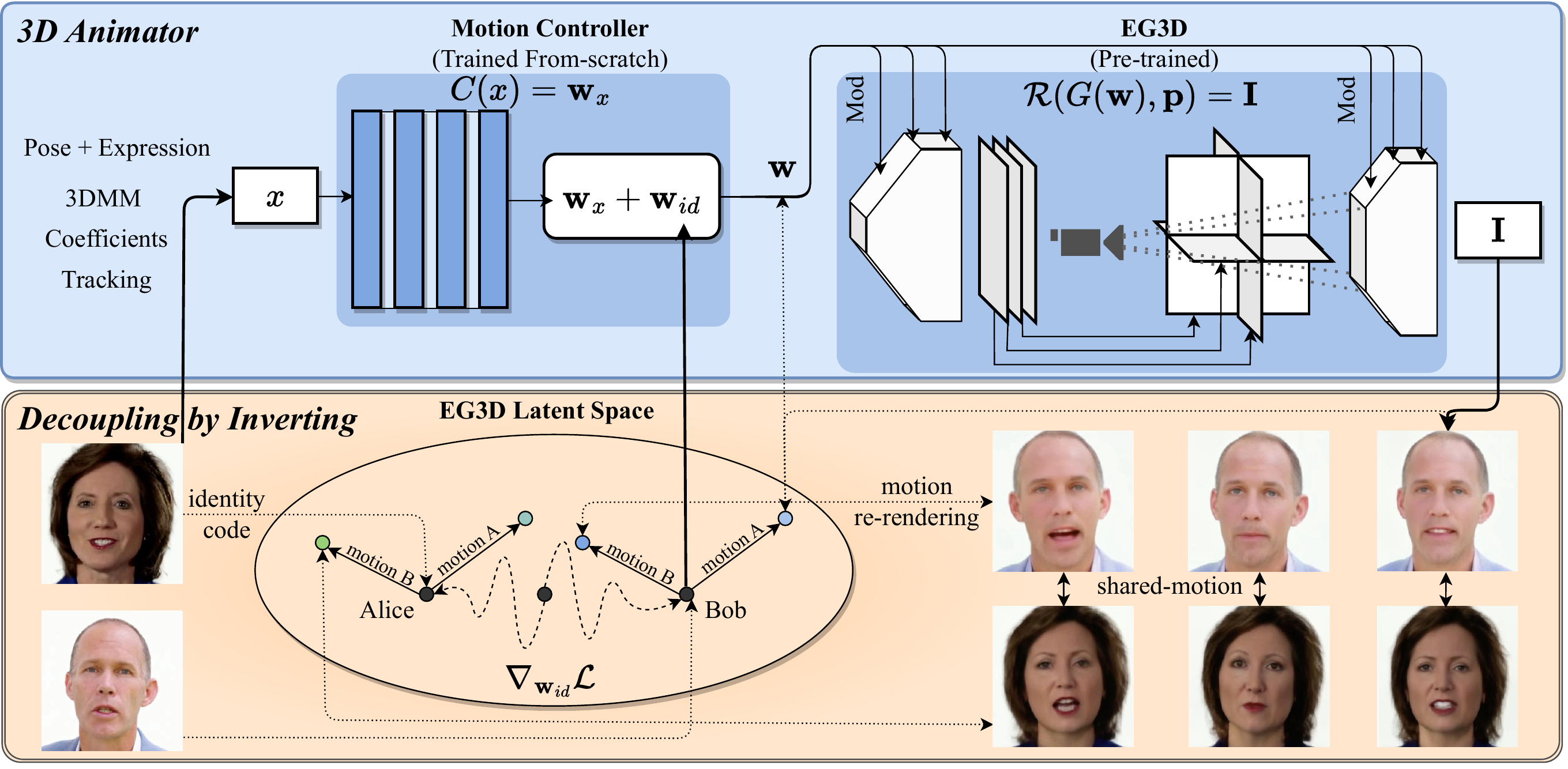}

   \caption{\textbf{Overview of OTAvatar.} OTAvatar contains a 3D face animator network and a latent code decoupling strategy, namely decoupling-by-inverting. The 3D face animator is composed of a pre-trained 3D face generator\cite{Chan2022} and a motion controller module. The decoupling-by-inverting algorithm is an optimization-based image inversion that can decouple the latent code $\mathbf{w}$ into identity code $\mathbf{w}_{id}$ and motion code $\mathbf{w}_x$. When the model is well-trained,  the motion-free identity code can be inferred from a single reference image, and an avatar of an unseen subject can be constructed. The identity code can be integrated with any other motion code predicted by the controller to animate the identity with desired motion.}
   \label{fig:overview of OTAvatar}
\end{figure*}

\section{Related Works}


\subsection{Talking Face}


    Regarding the driving signal, the talking face methods can be roughly divided into three categories: audio-driven, image-driven and coefficients-driven. Our approach is most related to coefficients-driven methods, which use either facial landmarks or 3DMM coefficients to represent motion. Facial landmarks involve identity and expression information and is challenging to transfer motion across different subjects. 3DMM coefficients disentangle the identity, expression and pose, and thus they are good driven signals to drive different faces. 
    
    Thies et al.~\cite{thies2016face2face} used 3D rendering to maintain the shape and illumination attributes when transferring expression and refining mouth details with a mouth retrieval algorithm. Geng et al.~\cite{geng20193d} used 3DMM to render a given subject with different expressions, followed by a neural network to refine the detail and harmonize with the background. These methods cannot change the face pose. Ren et al.~\cite{ren2021pirenderer} mapped 3DMM expression and pose to a high-dimensional motion representation, then predicted the dense image flow to reenact faces via image warping. Doukas et al.~\cite{doukas2021headgan} used additional 3DMM mesh fitting results to assist the dense image flow prediction. Yin et al.~\cite{yin2022styleheat} integrated pre-trained StyleGAN\cite{chong2021stylegan} to generate high-resolution talking face prediction by warping low-resolution feature maps. These warping-based methods follow the two-stage workflow by first warping images and then refining facial detail, demonstrating superior performance for same identity reenactment. But in cross-identity reenactment, especially when there is a large pose variation against the source portrait, the animation tends to collapse, since image warping is merely trained to stimulate 3D motion in the 2D space. Our work animates face avatars using 3D rendering and explicitly ensures 3D consistency to handle large pose variations.

\subsection{Volume Rendering}

Volume rendering is a 3D rendering strategy and has demonstrated its success in novel view synthesis. Existing methods either use volume~\cite{fernando2004gpu, Lombardi:2019} to explicitly represent 3D space or implicitly store the 3D scene in the MLP network~\cite{mildenhall2021nerf}. These methods are introduced to animate 3D-consistent portraits. Yu et al.~\cite{yu2021pixelnerf} and Raj et al.~\cite{raj2021pva} retrieved the spatial information from sparse support views to render portraits. On the other side, The 3D face generative model can reconstruct a given identity using GAN Inversion. These models vary from pure MLP architecture~\cite{chan2021pi}, to low-resolution neural rendering followed by an up-sampling strategy~\cite{schwarz2020graf, niemeyer2021giraffe}, and to using more efficient 3D presentation~\cite{Chan2022}. All these methods only support rendering static portraits. To animate the avatar with controllable motion, AD-NeRF\cite{guo2021ad} and NeRFace\cite{gafni2021dynamic} introduce either the audio feature or 3DMM expression coefficients to the NeRF model to render dynamic heads. Nerfies\cite{park2021nerfies} employs a deformation field conditioned on spatial points to represent the face motion of different frames. HyperNeRF\cite{park2021hypernerf} adds an ambient slicing network to tackle the topological drawback of Nerfies. Though these methods have achieved motion control and view consistency for face avatars, they all overfit each NeRF model to one subject, thus lack generalizability.  HeadNeRF~\cite{hong2022headnerf} is trained on both multi-view and frontal face datasets and can be generalized to multiple identities and support motion control using 3DMM coefficients, but its rendering quality is not satisfactory compared with the given portrait, and the motion control is unnatural with random jitters. In contrast, our method can reconstruct photo-realistic face avatars in one shot and animate natural talking motion. 

\subsection{Generative Prior}
\textcolor{black}{
Many methods rely on pre-trained generative models. Employing a generative model as the prior has the following advantages. First, it provides rich and diverse facial priors, such as texture, color, shape and pose, which can help to restore realistic and faithful facial details. For instance, Yang et al.\cite{Yang_2021_CVPR} leveraged a pre-trained GAN model as a prior decoder for blind face restoration. Second, it also enables one-shot image manipulation both inside and outside of the domain. Well-designed schemes for disentanglement are necessary to provide acceptable performance on fine-grained out-of-domain image manipulation. Zhang et al.\cite{zhang2022generalized} suggested separating the attributes into global style—like texture and color—and structure—like shape and pose, and performing domain-adaptive generation by transferring the decoupled style attributes. The image manipulation also includes synthesizing human talking videos, Ivan et al. \cite{skorokhodov2022stylegan} modeled the temporal dynamics of video frames through the latent code disentanglement. Unlike other talking head generation methods\cite{yin2022styleheat,siarohin2019first}, it only synthesizes center-aligned faces. The recently presented 3D GANs have been popularized for their ability to generate 3D-consistent photo-realistic human faces with explicit pose control. Katja et.al \cite{schwarz2020graf} introduced 3D scene volume rendering without resorting to computationally demanding voxel-based representation, and the model is trained from unposed 2D photos in an adversarial manner. Chan et al.\cite{Chan2022} proposed to generate high-quality geometry and multi-view consistent images from 2D photos through a hybrid explicit-implicit network architecture based on StyleGAN2\cite{karras2020analyzing}. Despite the fact that 3D GANs' explicit pose control and 3D-consistent generation are appealing for talking face synthesis, no previous work has studied their application to talking face avatars. 
}

\section{Method}
Talking face avatar aims to synthesize face images with controllable expression and pose. In this paper, we investigate the possibility of endowing volume rendering with the ability to 1) build a faithful identity representation in one shot, 2) enable natural motion control, and 3) achieve real-time inference speed.
 Our framework takes an identity code and a motion signal as input and generates a tri-plane-based~\cite{Chan2022} volume through a CNN architecture. 
 By performing volume rendering on the tri-plane representation, where each point is projected on the three feature planes and the sampled features are fused to occupancy and color, a face image at a given camera view can be generated as:


\begin{equation}
\label{equ-briefview}
\mathbf{I}(\mathbf{p}) = \mathcal{R}(\emph{G}(\mathbf{w}_{id},\mathbf{x};\Theta), \mathbf{p}),
\end{equation}
where $\mathbf{w}_{id}$ is the identity code, $\mathbf{x}$ is the motion signal, $\emph{G}(\cdot,\cdot;\Theta)$ is a face volume generation network with $\Theta$ as its network weights to reconstruct an encoded portrait in three orthogonal feature planes, $\mathcal{R}(\mathbf{V}, \mathbf{p})$ is the volume rendering with $\mathbf{V}$ as the volume and $\mathbf{p}$ as the camera view, and $\mathbf{I}(\mathbf{p}) \in \mathbb{R}^{3 \times H \times W}$ is the rendered result. In our implementation, the identity code $\mathbf{w}_{id}$ can be extracted by optimizing on a single reference image using our proposed decoupling-by-inverting strategy (Sec.~\ref{chaper:Latent code generation}), achieving one-shot avatar reconstruction. The motion signal $\mathbf{x}$ is 3DMM coefficients for flexible control. We call $\emph{G}(\cdot,\cdot;\Theta)$ as \textbf{3D Animator} because of its ability to animate an avatar with desired motion. Compared with 2D methods, we construct a neural 3D space to promise multi-view consistency. Compared with NeRF, we build a 3D space in the propagation of a CNN, and directly sample features on the tri-plane representation, saving the MLP computation on each sampled point.

\subsection{3D Animator Network Structure}
\label{chapter:Tri-plane 3D Representation and Rendering}

\textbf{Tri-plane volume representation.} The output of the 3D animator network $\emph{G}(\cdot,\cdot;\Theta)$ in Eqn.~\ref{equ-briefview} is a tri-plane volume representation, which is composed of three feature planes:
\begin{equation}
\label{equ-triplane}
\mathbf{V}_{tri} = (\mathbf{F}_{xy}, \mathbf{F}_{xz}, \mathbf{F}_{yz}) =  \emph{G}(\mathbf{w}_{id},\mathbf{x};\Theta),
\end{equation}
where $\mathbf{F}_{xy}$, $\mathbf{F}_{xz}$, and $\mathbf{F}_{yz}$ are three axis-aligned orthogonal feature maps in the 3D space, implicitly forming the volume $\mathbf{V}_{tri}$. When performing volume rendering, for each queried point $(x,y,z)$, we project it onto each of three feature maps and retrieve the corresponding features $(\mathbf{F}_{xy}(x,y), \mathbf{F}_{xz}(x,z), \mathbf{F}_{yz}(y,z))$. The summation of features is sent to a lightweight MLP to decode color and opacity. Compared to fully implicit MLP architectures\cite{guo2021ad, zheng2022imavatar, chan2021pi} or volume representation\cite{Lombardi:2019}, we can efficiently regress the three 2D feature maps through a CNN architecture. In our implementation, 
$\emph{G}$ is a deconvolutional network~\cite{chong2021stylegan} that outputs three $256 \times 256 \times 32$ feature maps. 

\textbf{Animator structure.} Towards animating the face avatar of a given portrait, an intuitive solution is representing $\mathbf{w}_{id}$ with a pre-defined feature and training $\emph{G}(\cdot)$ from scratch. However, we find that the model does not work well due to the sub-optimal $\mathbf{w}_{id}$. For example, when employing 3DMM shape coefficients~\cite{deng2019accurate} and face recognition features\cite{deng2019arcface}, the identity information is not preserved in the rendering results because $\mathbf{w}_{id}$ does not encode face appearances well. These models are trained on cropped facial images which contain no torso and hairstyle. Therefore, $\mathbf{w}_{id}$ losses such information that is necessary when animating portraits. The network even collapses if we make $\mathbf{w}_{id}$ learnable. 

In this work, we employ a two-phase strategy to achieve one-shot avatar reconstruction: 1) building a 3D face generator, and 2) making the generator controllable. 
We build our 3D face generator on a pre-trained EG3D\cite{Chan2022} network, which incorporates tri-plane representation for efficient 3D face generation:
\begin{equation}
\label{equ-eg3d}
\mathbf{V}_{tri} = \emph{G}_{eg}(\mathbf{w};\Theta_{eg}),
\end{equation}
where $\mathbf{w} $ is an uninterpretable latent code \textcolor{black}{drawn from the latent space of the generator $\emph{G}_{eg}$. The latent space is determined by the generator weight $\Theta_{eg}$}.
To control the latent code $\mathbf{w}$ by the given identity represented by $\mathbf{w}_{id}$ and motion signal $\mathbf{x}$, we propose a motion controller module $\emph{C}$ parameterized by $\Theta_{c}$, which maps motion signal $\mathbf{x}$ to the motion code $\mathbf{w}_x$, such that:
\begin{equation}
\label{equ-controller}
\mathbf{w} = \mathbf{w}_{id} + \mathbf{w}_{x} = \mathbf{w}_{id} + \emph{C}(\mathbf{x}; \Theta_{c}).
\end{equation}
Injecting Eqn.~\ref{equ-controller} to Eqn.~\ref{equ-eg3d}  , we have the structure of the 3D animator:
\begin{equation}
\begin{aligned}
\label{equ-3d-animator-net}
\emph{G}(\mathbf{w}_{id},\mathbf{x};\Theta) &= \emph{G}_{eg}(\mathbf{w}_{id} + \emph{C}(\mathbf{x}; \Theta_{c}); \Theta_{eg}), \\
\Theta &= \Theta_{eg} \cup  \Theta_{c}.
\end{aligned}
\end{equation}
In our implementation, the motion signal $\mathbf{x}$ is constructed by 3DMM pose and expression coefficients~\cite{deng2019accurate}, which are more flexible compared with image-based driving signals~\cite{siarohin2019first, siarohin2021motion, wang2022latent}, especially in cross-identity driving. 

\textbf{Controller structure.} We use the concatenation of 3DMM pose and expression coefficients to represent motion $\mathbf{x}$. Practically, we use the adjacent coefficients of pose and expression within a particular radius to represent the motion signal of the current frame, firstly the weighted summation of the coefficients across the temporal dimension is conducted to avoid noises, which is achieved by three 1D convolution layers\cite{ren2021pirenderer, yin2022styleheat}; secondly, a five-layer MLP is employed to transform the weighted summation into a motion feature; finally, a codebook with learnable orthogonal bases~\cite{wang2022latent} is built, on which each motion feature is projected to get the final motion code $\mathbf{w}_x$. The aforementioned operation is summarized as follows:
\begin{equation}
\begin{aligned}
    \mathbf{w}_x = \emph{C}(\mathbf{x}; \Theta_{c}) = & ~\mathbf{D} * ~\emph{F}_{M}(\emph{F}_{T}(\mathbf{x}; \Theta_{T}); \Theta_{M}) , 
    \label{equ-controller-2}
\end{aligned}
\end{equation}
where $\emph{F}_{T}$ parameterized by $\Theta_{T}$ is the temporal smoothing network of 1D convolution layers, $\emph{F}_{M}$ parameterized by $\Theta_{M}$ is the five-layer MLP which projects motion to the magnitudes of bases contained in the codebook $\mathbf{D}$. It can be seen that the magnitude projection parameters, temporal smoothing weights, and the codebook are the parameters of the controller $\emph{C}$ to be trained, that is:
\begin{equation}
\Theta_{c}=\mathbf{D}  \cup  \Theta_{T} \cup \Theta_{M}.
\end{equation}

\subsection{Controller Training}
\label{sec:alternative manner}

Given a couple of source frame $\mathbf{I}_s$ and driving frame $\mathbf{I}_d$, along with their motion signals $\mathbf{x}_s, \mathbf{x}_d$ and camera views $\mathbf{p}_s, \mathbf{p}_d$, we perform the dual-objective optimization:
\begin{equation}
\label{equ-controller-training}
\begin{aligned}
  \mathbf{w}^*_{id} &= \underset{\mathbf{w}_{id}}{\arg \min } (  \mathcal{L}_s + \mathcal{L}_d ),\\
  \mathcal{L}_s = \mathcal{L}(I_s, \hat{I}_s) &= \mathcal{L}(I_s, \mathcal{R}(\emph{G}(\mathbf{w}_{id}, \mathbf{x}_s; \Theta), \mathbf{p}_s), \\
  \mathcal{L}_d = \mathcal{L}(I_d, \hat{I}_d) &= \mathcal{L}(I_d, \mathcal{R}(\emph{G}(\mathbf{w}_{id}, \mathbf{x}_d; \Theta), \mathbf{p}_d).
\end{aligned}    
\end{equation}
It is implemented to estimate the identity code $\mathbf{w}_{id}$ shared between $\mathbf{I}_s$ and $\mathbf{I}_d$.
$\emph{G}$, $\mathcal{R}$, and $\mathcal{L}$ indicate 3D face generator, volume rendering, and loss function, respectively. The 3D face animator $G$ is parameterized by $\Theta = \Theta_{eg} \cup \Theta_{c}$. $\Theta_{eg}$ includes the parameters of the pre-trained 3D face generator, and $\Theta_{c}$ includes the parameters of the controller which is the focus of our training scheme. 


\textbf{Training scheme}. The purpose of controller training is to decouple the identity code and motion code from the latent code of the generator, making identity and motion replaceable in our generalized avatars. To this end, we propose the \textit{decoupling-by-inverting} training strategy, which is generally an alternation of identity optimization and controller training at different steps. Specifically, we first freeze $\Theta_{c}$ and perform $N_{id}$ steps of back-propagation on $\mathbf{w}_{id}$. 
Then we freeze $\mathbf{w}_{id}$ and perform $N_{mo}$ steps of back-propagation on $\Theta_{c}$ so that the controller learns to add the motion code on top of the identity code.
In our implementation, $N_{id}=90$ and $N_{mo}=10$. 
After $N = N_{id} + N_{mo}$ steps, we slightly finetune $\Theta_{eg}$ with a learning rate of $10^{-4}$ to make the rendering fit the driving signal better. A brief pseudo-code of the whole scheme is provided in Alg.~\ref{alg:Inverting while Decoupling Speed-Up Algorithm}.

\subsection{One-shot Avatar Construction}
\label{chaper:Latent code generation}
In the inference stage, given a single portrait image $\mathbf{I}_{r}$, its 3DMM coefficients $\mathbf{x}_{r}$, and camera view $\mathbf{p}_{r}$ as the reference of avatar construction, we can estimate the identity code $\mathbf{w}_{r}$ of the reference portrait via: 
\begin{equation}
  \mathbf{w}_{r}=\underset{\mathbf{w}_{id}}{\arg \min }  \mathcal{L}(\mathbf{I}_r, 
  \mathcal{R}(\emph{G}(\mathbf{w}_{id}, \mathbf{x}_r), \mathbf{p}_r).
  \label{equ-gan inversion with control test}
\end{equation}
Afterwards, we can animate the identity with given driving motion $\mathbf{x}_d$ and camera view $\mathbf{p}_d$, either from the same or different subject, through the following operation: 
\begin{equation}
    \label{equ-same/cross reenactment}
    \mathbf{I}(\mathbf{w}_{r}, \mathbf{x}_d, \mathbf{p}_d) = \mathcal{R}(\emph{G}(\mathbf{w}_{r}, \mathbf{x}_d), \mathbf{p}_d),
\end{equation}
where $\mathbf{I}(\mathbf{w}_{r}, \mathbf{c}_d, \mathbf{p}_d)$ has the identity estimated as $\mathbf{I}_r$, and is animated with the motion $\mathbf{c}_d$ and camera view $\mathbf{p}_d$. We name Eqn.~\ref{equ-gan inversion with control test} as the \textit{decoupling-by-inverting} inference strategy since it can estimate decoupled identity code from latent code in the iterative optimization-based GAN Inversion, and can be used to render different motions. 



\subsection{Loss Terms}
\label{Chapter:Loss Terms}

The loss $\mathcal{L}(\mathbf{I}, \hat{\mathbf{I}})$ in Eqn.~\ref{equ-controller-training} and Eqn.~\ref{equ-gan inversion with control test} measures the error between the animated result $\hat{\mathbf{I}}$ and the ground truth $\mathbf{I}$. First, a pre-trained VGG-19 network is implemented to calculate the distance of multi-scale activation maps $\phi_i(\textbf{I})$ via:
\begin{equation}
\mathcal{L}_c=\sum_i\left\|\phi_i(\mathbf{I})-\phi_i(\hat{\mathbf{I}})\right\|_1,
\end{equation}
where $\phi_i$ is the $i$-th layer of VGG. Second, the distance between the Gram matrices $G_j^\phi$ constructed from the $j$-th activation maps $\phi_j$\cite{ren2021pirenderer} of $\mathbf{I}$ and $\hat{\mathbf{I}}$ is measured:
\begin{equation}
\mathcal{L}_s=\sum_j\left\|G_j^\phi(\mathbf{I})-G_j^\phi(\hat{\mathbf{I}})\right\|_1.
\end{equation}
Third, we utilize the L1 distance performed on the eyes and mouth regions to supervise the expression detail:
\begin{equation}
\mathcal{L}_m=\sum_n\left\|R_n(\mathbf{I})-R_n(\hat{\mathbf{I}})\right\|_1,
\end{equation}
where $R_n$ is either the eyes or mouth region extracted using RoI-align on the bounding boxes calculated using landmarks. Finally, the ID Loss is performed to preserve the identity consistency:
\begin{equation}
\mathcal{L}_{i d}=1-\cos (E(\mathbf{I}), E(\hat{\mathbf{I}})),
\end{equation}
where $E$ is the Arcface face recognition model \cite{deng2019arcface}. 

The animation loss $\mathcal{L}(\mathbf{I}, \hat{\mathbf{I}})$ is a weighted summation of the above loss terms and is implemented in both training and test-time one-shot avatar construction. Additionally, in the training, we maintain the stability of tri-plane representation with monotonic and TV loss\cite{Chan2022}, and $\mathcal{L}_1$ loss on $\Theta_{c}$ to avoid the latent code being out of the latent distribution of $\emph{G}_{eg}$. 

\section{Experiments}

\subsection{Experiment Setting}

In this section, we first describe the implementation detail of our proposed method, the dataset used, and the baselines of our work. Then we compare our proposed method with previous 2D and 3D methods on motion controllability and multi-view consistency.

\textbf{Implementation details}. 
We use the off-the-shelf 3D face reconstruction model to extract the expression and pose coefficients of 3DMM\cite{deng2019accurate}. The motion of any timestamp is represented by the window of adjacent 27 frames of expression and pose coefficients. In practice, we find that the extracted head poses contain too much noises to serve as the camera pose. Therefore, we  extract the camera pose using the method in~\cite{guo2021ad}. Our model is implemented in Pytorch using four A100 GPUs. The total batch size is 24, with six images per GPU. We use Adam optimizer with a 0.0001 learning rate to train the motion controller $\emph{C}$ and finetune the generator $\emph{G}$. Exponential Moving Average (EMA) is employed to update $\Theta_{c}$ in Alg.~\ref{alg:Inverting while Decoupling Speed-Up Algorithm} since it can stabilize the training. The model is trained for 2000 iterations, and each iteration contains $N = N_{id} + N_{mo} = 100$ steps to optimize the identity codes and train the motion controller. For each mini-batch of data, a new batch of latent codes are initialized and optimized using the Adam optimizer with a 0.01 learning rate. Both the 64$\times$64 resolution volume rendering results and 512$\times$512 super-resolution results are implemented to calculate the loss.  
\textcolor{black}{During training, the identity code $\mathbf{w}_{id}$ is optimized in $\mathcal{W}$ space. After being extended to the $\mathcal{W}^+$ space by channel-wisely repeated 14 times, it is combined with the motion code $\mathbf{w}_x$ and then fed into the face generator $\emph{G}$.}
It takes less than two days to train the network entirely.
\textcolor{black}{During inference, we first optimize the identity code in $\mathcal{W}$ space, then use another $N=100$ steps to finetune it in $\mathcal{W}^+$ space~\cite{abdal2019image2stylegan, abdal2020image2stylegan++}. }

\textbf{Dataset}. Our model is trained on the HDTF\cite{zhang2021flow} dataset, which contains frontal talking faces from 362 videos and over 300 subjects. Following \cite{yin2022styleheat}, we use the pre-processing step in \cite{siarohin2019first} and resize images to 512$\times$512 resolution. The test dataset contains 20 videos of subjects out of the training set. To evaluate the novel subject generalizability and controllability 
in the 3D-consistent fashion, we sample subjects from the Multiface\cite{wuu2022multiface} dataset, which is particularly proposed for 3D-consistent face avatars. 

\textbf{Baseline methods}. We mainly compare our method with previous works that incorporate 3DMM coefficients to reenact face avatars in either 2D/3D fashion and are generalizable to unseen identities. For 2D methods, PIRenderer~\cite{ren2021pirenderer} and StyleHEAT~\cite{yin2022styleheat} achieve SOTA performance on face reenactment.
For 3D methods, HeadNeRF~\cite{hong2022headnerf} is the most advanced volume rendering method that incorporates controllability and generalizability. \textcolor{black}{Beyond coefficients-based methods, the SOTA image-driven method FOMM~\cite{siarohin2019first} is also taken into comparison}.

\begin{figure}[ht]
  \centering
   \includegraphics[width=1.\linewidth]{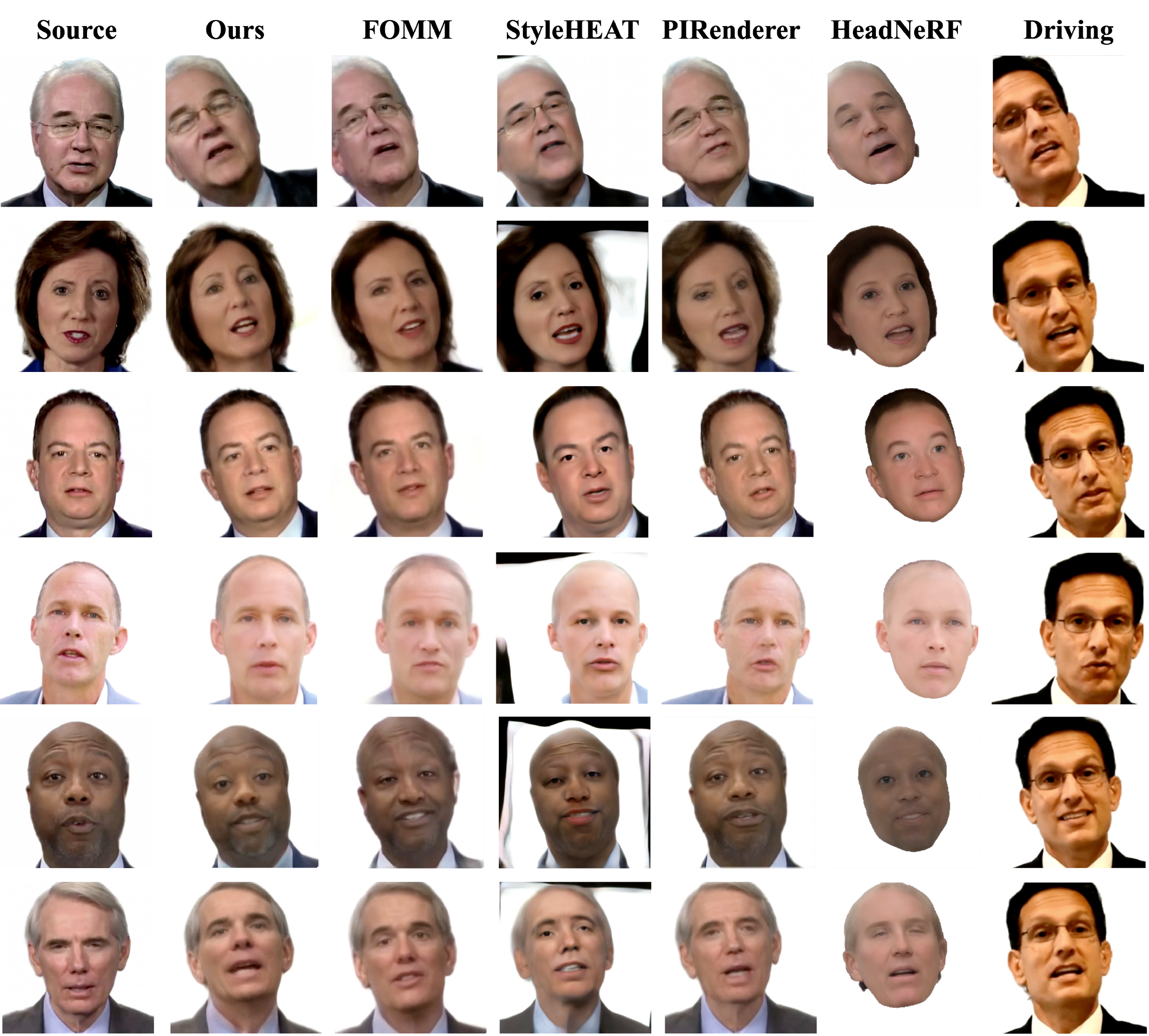}

   \caption{\textbf{Qualitative result for cross-identity reenactment}. Examples are sampled from the HDTF dataset~\cite{zhang2021flow}. Both source and driving subjects are not included in the training set. }
    \label{fig:hdtf comparison}
\end{figure}

\begin{figure*}[ht]
  \centering
   \includegraphics[width=1\linewidth]{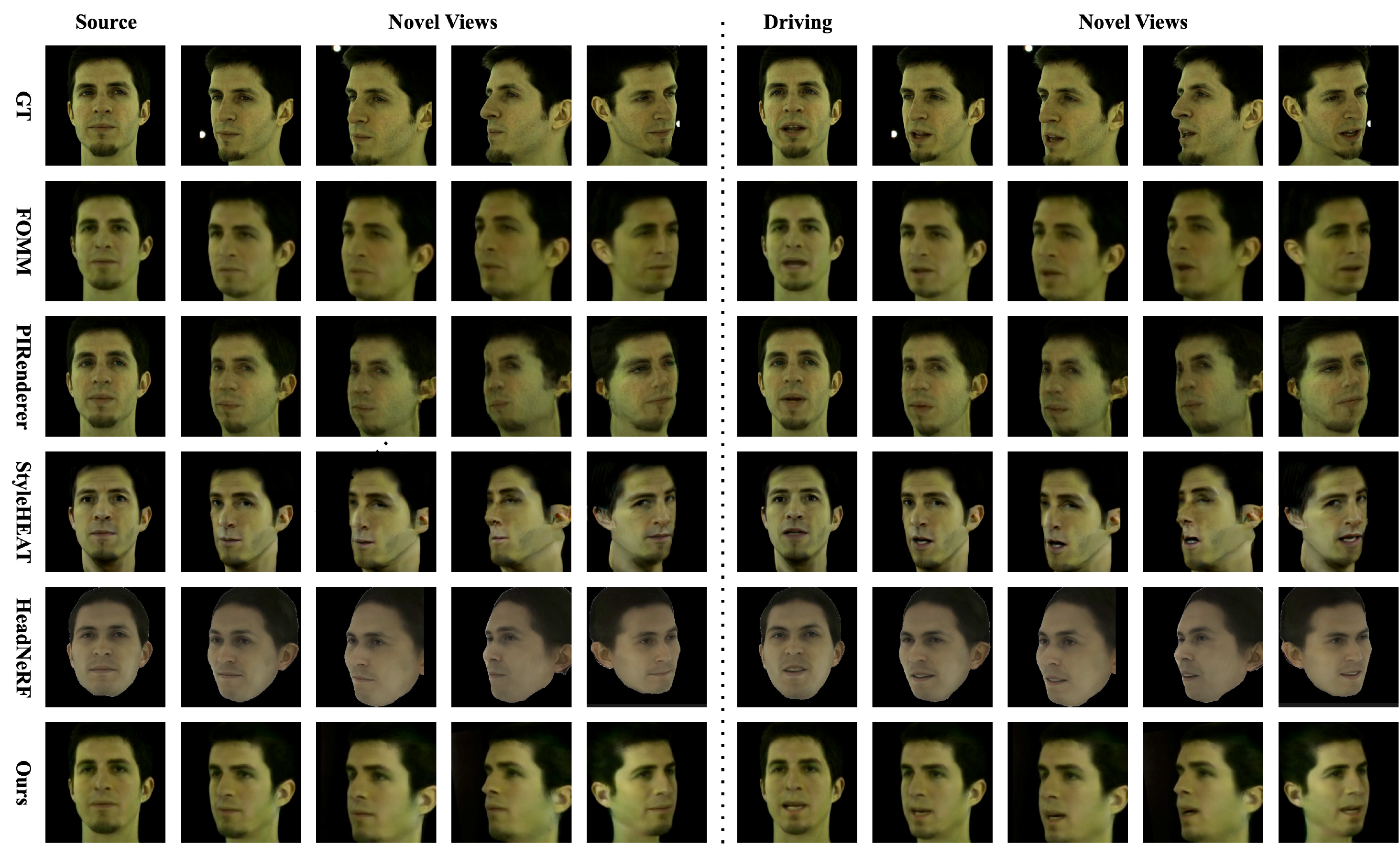}
   \caption{\textbf{Qualitative result for multi-view reenactment}. Examples are sampled from the multi-view dataset~\cite{wuu2022multiface}. All methods use the first frame of the frontal-view portrait to extract the identity feature, and take the expressions of sequential frames and poses of different camera views to generate the talking face. Note that this subject is not included in the training set of any methods. }
    \label{fig:multiface comparison}
    \vspace{+2mm}
\end{figure*}

\textbf{Evaluation metrics}. 
We adopt peak-signal-to-noise ratio (PSNR), structural similarity index measure (SSIM)~\cite{wang2004image}, and learned perceptual image patch similarity (LPIPS)~\cite{zhang2018unreasonable} to evaluate the visual quality. To measure the realism of the synthesized results and the identity preservation, we use frechet inception distance (FID)~\cite{heusel2017gans}  and cosine similarity of identity embedding (CSIM)~\cite{deng2019arcface} based on ArcFace model, respectively. Besides, the average expression distance (AED), average pose distance (APD), and average keypoint distance (AKD) are employed to evaluate the facial expression and pose.

\begin{table*}[ht]
\centering
\resizebox{\textwidth}{!}{%
\begin{tabular}{@{}c|cccccccc|ccccc@{}}
\toprule
 &
  \multicolumn{8}{c|}{Multi-View Reenactment} &
  \multicolumn{5}{c}{Cross-Identity Reenactment} \\ \cmidrule(l){2-14} 
 &
  PSNR $\uparrow$ &
  SSIM $\uparrow$ &
  CSIM $\uparrow$ &
  AED $\downarrow$ &
  APD $\downarrow$ &
  AKD $\downarrow$ &
  LIPIPS $\downarrow$ &
  FID $\downarrow$ &
  CSIM  $\uparrow$ &
  AED $\downarrow$ &
  APD   $\downarrow$ &
  AKD $\downarrow$ &
  FID $\downarrow$ \\ \midrule
FOMM &
  20.75 &
  0.639 &
  0.505 &
  2.004 &
  0.545 &
  5.052 &
  0.308 &
  101.6 &
  0.672 &
  3.196 &
  0.500 &
  4.198 &
  113.2 \\
PIRenderer &
  20.04 &
  0.586 &
  0.493 &
  2.203 &
  0.680 &
  6.566 &
  0.299 &
  \textbf{100.6} &
  0.632 &
  3.018 &
  0.498 &
  4.977 &
  103.7 \\
StyleHEAT &
  20.03 &
  0.632 &
  0.387 &
  2.179 &
  0.472 &
  5.522 &
  \textbf{0.284} &
  123.8 &
  0.614 &
  2.860 &
  0.471 &
  3.592 &
  239.1 \\
HeadNeRF &
  17.60 &
  0.546 &
  0.239 &
  2.086 &
  0.776 &
  4.166 &
  0.367 &
  212.3 &
  0.282 &
  2.873 &
  0.567 &
  \textbf{3.465} &
  233.0 \\
Ours &
  \textbf{21.19} &
  \textbf{0.657} &
  \textbf{0.574} &
  \textbf{1.874} &
  \textbf{0.428} &
  \textbf{3.731} &
  0.288 &
  137.3 &
  \textbf{0.694} &
  \textbf{2.850} &
  \textbf{0.405} &
  4.307 &
  \textbf{101.8} \\ \bottomrule
\end{tabular}%
}
\caption{\textbf{Quantitative comparisons on multi-View reenactment and cross-identity reenactment.} }
\label{tab:quantative comparison}
\end{table*}

\subsection{Evaluation Result}
We evaluate OTAvatar on animating photo-realistic talking head videos and compare them with state-of-the-art models that support identity-generalized animation methods. In the HDTF~\cite{zhang2021flow} dataset, we examine the identity-motion disentanglement by transferring the motion of one subject to drive the other subjects, namely cross-identity reenactment. The employed motion is extracted from a video with large motion variation to evaluate the result in extreme conditions.
In the Multiface~\cite{wuu2022multiface} dataset, we evaluate the consistency of the animation in different views, namely multi-view reenactment. Note that none of the data in this dataset has been used in training our method and baselines. For each talking corpus, we choose the first frame of the frontal-view camera recording as the reference and enforce the network to animate the following frames in frontal view and other views. 

\textbf{Qualitative comparison}. Fig.~\ref{fig:hdtf comparison} shows the results of different methods on cross-identity generation. Compared to FOMM~\cite{siarohin2019first}, PIRenderer~\cite{ren2021pirenderer} and StyleHEAT~\cite{yin2022styleheat}, which use warping fields for reenacting faces, our method can handle extreme pose variation and maintain identity consistency. Compared to the 3D method of HeadNeRF, our model fully reconstructs the identity-specific detail and synthesizes more natural expression. The multi-view consistency results are shown in Fig.~\ref{fig:multiface comparison}. First, the 2D warping method suffers facial malformation, which becomes more serious in larger poses. Second, though HeadNeRF renders accurate head poses, the generated results have noticeable deterioration compared to the ground truth. Finally, the face avatar constructed by our method preserves identity details and multi-view consistency. 

\textbf{Quantitative comparison}. The quantitative comparison among competing methods is shown in Table~\ref{tab:quantative comparison}. We can see that the proposed method achieves the best performance in terms of most of the criteria. 

\textbf{Inference speed}. Table~\ref{tab:inference speed} lists the inference speed comparison among state-of-the-art 3D face avatar methods. By utilizing the pre-trained 3D face generator and the compact motion controller (0.8M parameters), our method achieves the highest inference speed, demonstrating its effectiveness.

\begin{table}[ht]
\centering
\setlength{\tabcolsep}{0.7cm}
\begin{tabular}{cc}
\toprule
Methods  & Frames Per Second $\uparrow$ \\ \midrule
IMAvatar\cite{zheng2022imavatar} & 0.03                   \\
ADNeRF\cite{guo2021ad} & 0.13                     \\
HeadNeRF\cite{hong2022headnerf} & 25                     \\
Ours     & \textbf{35}                     \\ \bottomrule
\end{tabular}%
\caption{\textbf{Quantitative comparison on the inference speed}. The comparison is conducted on an A100 GPU. }
\label{tab:inference speed}
\end{table}

\begin{figure}[ht]
  \centering
   \includegraphics[width=.7\linewidth, ]{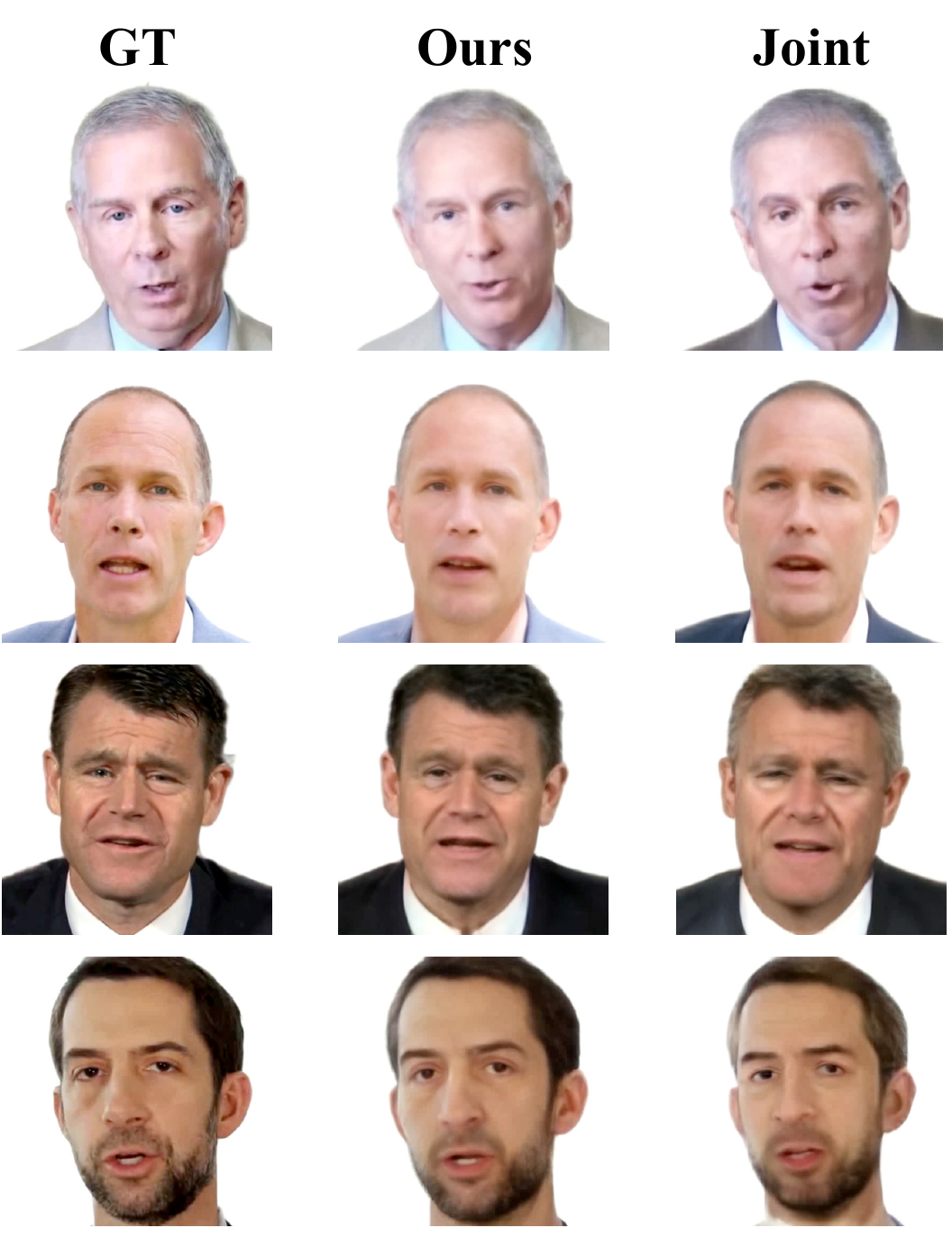}
    \caption{\textbf{Qualitative comparison of decoupling-by-inverting training and joint training}. Joint training cannot preserve identity information in one-shot avatar construction.}
    \label{fig:joint training comparison}
\end{figure}

\subsection{Ablation Study}

\textbf{Decoupling-by-inverting}. The controller is trained by the decoupling-by-inverting strategy described in Sec.~\ref{sec:alternative manner}, where the training alternates between $N_{id}$ steps of identity code optimization and $N_{mo}$ steps of controller parameters training. To validate the effectiveness of the alternative manner, we discard it and jointly train the controller parameters and the identity code. The qualitative comparison is in Fig.~\ref{fig:joint training comparison}. We see that the jointly trained model cannot preserve identity during the animation because the controller manages to overfit the identity information in the training set, and the identity information is not fully encoded in the identity code. The performance degradation caused by joint training is also shown in Table ~\ref{tab:loss ablation}. We also analyze the setting of identity optimization step $N_{id}$ and controller training step  $N_{mo}$, and list the results in Table~\ref{tab:updating strategy}. We observe that $N_{id} = 90, N_{mo} = 10$ achieves a balance of expression reconstruction (AED) and identity preserving (CSIM).

\textbf{Losses}. We also perform an ablation study on the loss functions, including the landmark region losses $\mathcal{L}_{m}$ and ID losses $\mathcal{L}_{id}$, as shown in Table~\ref{tab:loss ablation}. We can see that the absence of landmark region losses $\mathcal{L}_m$  causes a deterioration of expression reconstruction, measured by AED, while the absence of ID loss causes a performance drop in both expression and identity consistency.


\begin{table}[ht]
\centering
    \setlength{\tabcolsep}{0.5cm}
        \begin{tabular}{cccc}
            \toprule
            $\text{N}_\text{mo}$ & $\text{N}_\text{id}$ & AED $\downarrow$   & CSIM $\uparrow$  \\ 
            \midrule
            1   & 99 & 3.285 & \textbf{0.7200} \\
            5   & 95  & 3.178 & 0.716 \\
            10  & 90  & \textbf{2.850} & 0.694 \\
            20  & 80  & 3.231 & 0.592 \\
            \bottomrule
        \end{tabular}         
        \caption{\textbf{Ablation study on the decoupling-by-inverting hyperparameters.} Experiments are conducted under the different settings of $N_{mo}$ and $N_{id}$ with $N = N_{mo} + N_{id} = 100 $. }
        \label{tab:updating strategy}
\end{table}

\begin{table}[ht]
\centering
    \setlength{\tabcolsep}{0.5cm}
        \begin{tabular}{ccc}
        \toprule
              &        AED $\downarrow$   & CSIM $\uparrow$  \\ 
        \midrule
        ours   &     \textbf{2.850} & \textbf{0.694} \\
        joint &     3.009 & 0.689 \\
        w/o $\mathcal{L}_\text{m}$      & 3.101 & 0.687 \\
        w/o $\mathcal{L}_\text{id}$      & 3.164 & 0.592 \\
        w/o finetune & 3.145 & 0.687 \\ 
        \bottomrule
        \end{tabular}        
        \caption{\textbf{Ablation study on the \textcolor{black}{joint training}, loss terms, and finetuning}. 
        }
        \label{tab:loss ablation}
\end{table}

\textbf{3D generator finetuning}. When the controller training and identity code optimization are finished, we further slightly finetune the pre-trained EG3D~\cite{Chan2022} face generator with a learning rate of 0.0001. In Table~\ref{tab:loss ablation}, we show the results without finetuning. One can see that slightly finetuning the face generator in the final stage improves identity and expression reconstruction performance.

\section{Conclusion}
We proposed a novel framework of one-shot 3D-consistent talking face avatar, namely OTAvatar, using volume rendering. It jointly addressed the three challenges of face avatars, i.e., generalizability, controllability, and efficiency. For identity-generalized rendering, we implemented a pre-trained 3D face generator to reconstruct faces faithfully, given a single portrait reference. For motion control, we proposed a motion controller module, which predicts the motion code conditioned on 3DMM coefficients. For efficiency, benefiting from the compact architecture of both the 3D generator and controller, our model can animate avatars at a high speed. Besides, we proposed a decoupling-by-inverting approach, which is both a training scheme and a test-time disentangle strategy that decouples the latent code into identity and motion codes via GAN inversion, so that one can animate the avatar with different motions using the motion codes predicted by the controller. 
Comprehensive experiments were conducted to prove the advantage of our proposed framework.


\section*{Acknowledgement}
This work is supported in part by the Chinese National Natural Science Foundation Projects $\#62176256$, $\#62276254$, and the InnoHK program.



{\small
\bibliographystyle{ieee_fullname}
\bibliography{egbib}
}

\clearpage

\section{Supplementary Material}

    

\subsection{Pesudo Code}

The pseudo-code for our suggested inverting-by-decoupling training scheme is appended in this section.

\begin{algorithm}[ht]

    \caption{Training Scheme of Decoupling-by-Inverting}
    \label{alg:Inverting while Decoupling Speed-Up Algorithm}


    \KwIn{Talking Face Dataset $\mathcal{D}$, \\
    3D face animator $\emph{G}(\cdot, \cdot ; \Theta) = \emph{G}_{eg}(\cdot + \emph{C}(\cdot; \Theta_{c});\Theta_{eg})$ \\
    
    where $\Theta = \Theta_{eg} \cup \Theta_{c}$ \\ 


    }
        
    $\Theta_{c} \gets$ random initialization \\  
    \For{$i \gets 1$ \KwTo $T$}{        
        \tcc{collect source and target data point} 
        $\mathcal{V} \gets$ random video clip sampled from $\mathcal{D}$ \\
        $\mathbf{I}_s , \mathbf{x}_s, \mathbf{p}_s \gets$ data of random frame sampled from $\mathcal{V}$ \\
        $\mathbf{I}_d , \mathbf{x}_d, \mathbf{p}_d \gets$ data of random frame sampled from $\mathcal{V}$

        $\mathbf{w}_{id} \gets \mathbf{w}_{avg}$ \tcp*[f]{initialize $\mathbf{w}_{id}$ using average} \;
        
        $\theta_{c} \gets \Theta_{c}$  \tcp*[f]{initialize $\theta_{c}$ using EMA weights} \; 
        
        $\theta \gets \Theta_{eg} \cup \theta_{c}$ \tcp*[f]{assemble \emph{G} with trainable $\theta_{c}$} \;
        
        \For{$n \gets 1$ \KwTo $N_{id} + N_{mo}$}{
            
            \tcc{calculate optimization objectives}
            

            $\mathcal{L}_s \gets \mathcal{L}(\mathbf{I}_s, \mathcal{R}(\emph{G}(\mathbf{w}_{id}, \mathbf{x}_s ; \theta), \mathbf{p}_s))$ \label{alg:loss_propogation_s}

            $\mathcal{L}_d \gets \mathcal{L}(\mathbf{I}_d, \mathcal{R}(\emph{G}(\mathbf{w}_{id}, \mathbf{x}_d ; \theta), \mathbf{p}_d))$ \label{alg:loss_propogation_d}
            
            \eIf{$n < N_{id}$}{
                \tcc{optimize identity code}
                Update $\mathbf{w}_{id}$ using $\nabla_{\mathbf{w}_{id}} (\mathcal{L}_s + \mathcal{L}_d)$
            }{
                \tcc{train motion calibration}
                Update $\theta_{c}$ using $\nabla_{\theta_{c}} (\mathcal{L}_s + \mathcal{L}_d)$
            }
        }
        Calculate $\mathcal{L}_s$, $\mathcal{L}_t$ using line.\ref{alg:loss_propogation_s}, line.\ref{alg:loss_propogation_d}
        
        Finetune $\Theta_{eg}$ on $\mathcal{L}_s + \mathcal{L}_t$ \label{alg:fine-tune G}
        
        $\Theta_{ca}  \gets \beta \Theta_{c} + (1 - \beta) \theta_{c}  $  \tcp*[f]{Update EMA weights} \; 
    }
 
\end{algorithm}

\subsection{Identity code interpolation}

\begin{figure*}[t]
  \centering
   \includegraphics[width=0.8\textwidth, ]{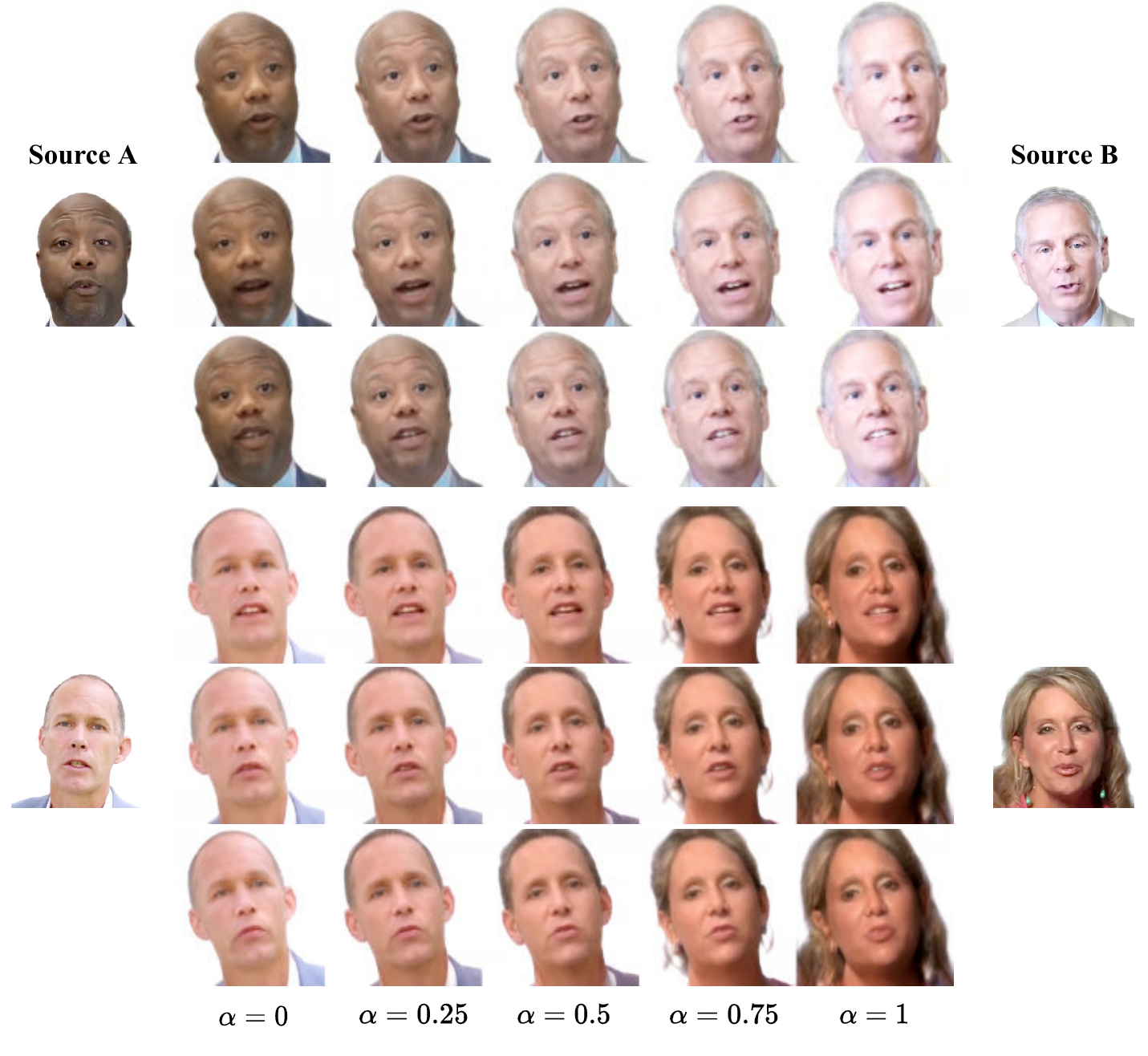}
    \caption{\textbf{The interpolation results of identity code}. Our model can generate smooth-varying identity attribute beyond motion control.}
    \label{fig:identity interpolation}
\end{figure*}

In this paper, we use a motion controller $\emph{C}$ and decoupling-by-inverting strategy to disentangle the latent code of face generator to identity code $\mathbf{w}_{id}$ and motion code $\mathbf{w}_{x}$. In this section, we examine the disentanglement by performing the task of identity interpolation. The interpolated identity is generated by:

 \begin{figure}[t]
  \centering
   \includegraphics[width=1.1\linewidth, ]{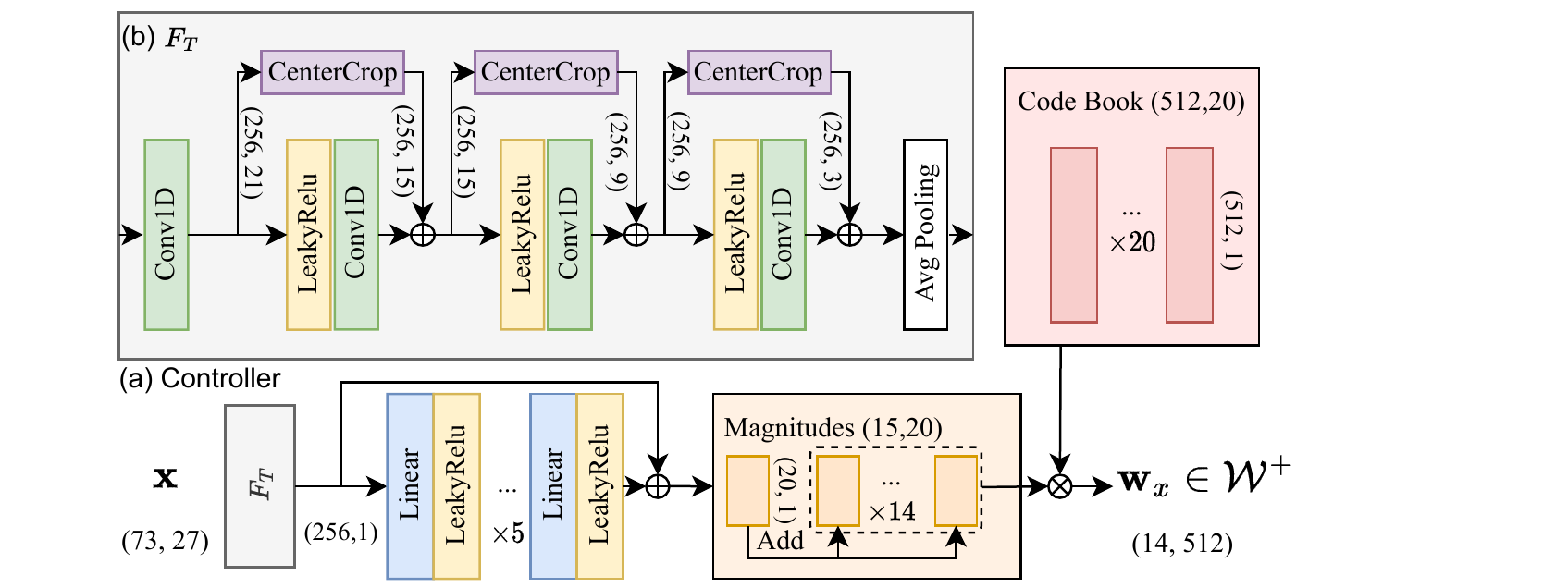}
    \caption{\textbf{Controller architecture}. We show the details of the architecture of \emph{C} in (a) and the sub-module $\emph{F}_T$ in (b). The controller takes as input 3DMM coefficients and output motion code in $\mathcal{W}^+$ space\cite{xia2022gan}.}
    \label{fig:controller}
\end{figure}

\begin{equation}
\label{equ-identity interpolation}
\mathbf{w}^{'}_{id} = \alpha \mathbf{w}_{id\_a} + (1 - \alpha) \mathbf{w}_{id\_b}
\end{equation}
where $\mathbf{w}_{id\_a}$ and $\mathbf{w}_{id\_b}$ are the identity codes estimated from two source reference images using decoupling-by-inverting strategy. The interpolated latent code $\mathbf{w}^{'}_{id}$ is then combined with any motion and generates animations. The animation result in shown in Fig.~\ref{fig:identity interpolation}. It shows that our model can animate consistent motion with smooth-varying identity attributes. And it validates that our method achieves the disentanglement of motion and identity in the latent space of the pre-trained face generator. The animation result is also appended in the supplementary video.

\begin{table}[ht]
	\centering
	\begin{tabular}{cccccc}
	\hline
& CSIM  & AED   & APD   & AKD   & FID   \\
\hline
$\mathbf{w}_x \in \mathcal{W}$ & \textbf{0.719} & 3.352 & 0.453 & 4.783 & 104.6 \\
w/o code book                                   & 0.662 & 3.342 & 0.457 & 4.974 & 103.2 \\
Ours                                            & 0.694 & \textbf{2.850} & \textbf{0.405} & \textbf{4.307} & \textbf{101.8} \\ 
    \hline
    \end{tabular}
\caption{\textbf{Ablation study on the controller architecture}. Experiments are conducted on the cross-identity reenactment.}
\label{tab:ablation_on_controller}
\end{table}

\subsection{More Qualitative Comparison}
    
\begin{figure*}[t]
  \centering
   \includegraphics[width=0.9\linewidth, ]{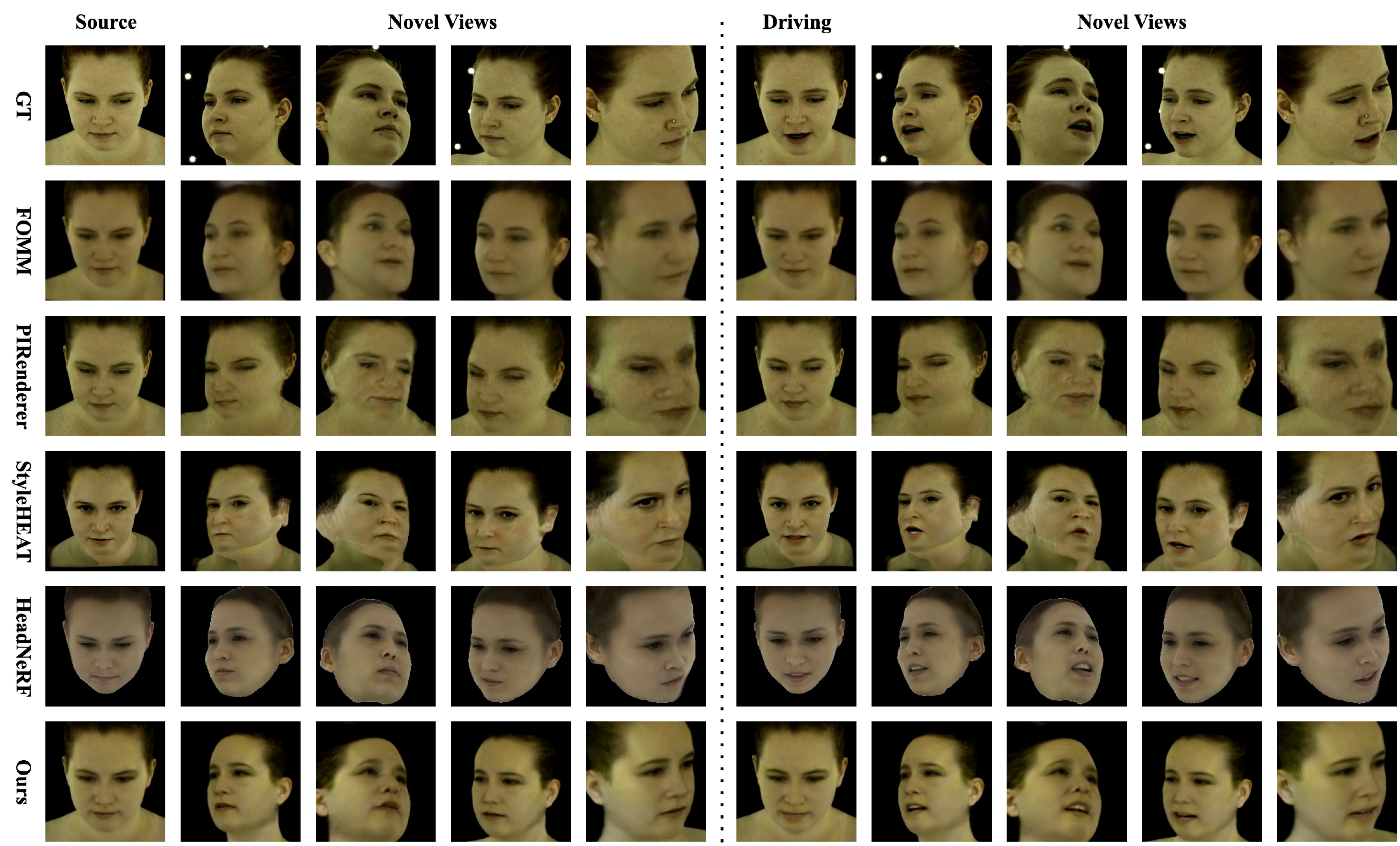}
    \caption{\textbf{3D Consistency on multi-view dataset}. We demonstrate additional visualization comparison on the Multiface dataset~\cite{wuu2022multiface}, with more drastic camera view variation. All methods use the source frame to extract the identity feature, then extract 3DMM coefficients of pose and expression from the driving frame to generate the talking face. This subject is not included in the training set of any methods.}
    \label{fig:multiface_woman}
\end{figure*}

\textbf{Multi-view dataset}. We show more qualitative comparison result on the multi-view stereo dataset Multiface~\cite{wuu2022multiface} in Fig.~\ref{fig:multiface_woman}. To further compare the robustness against the pose variation, we choose an overhead view as the single reference. 

\begin{figure*}[t]
  \centering
   \includegraphics[width=1\linewidth, ]{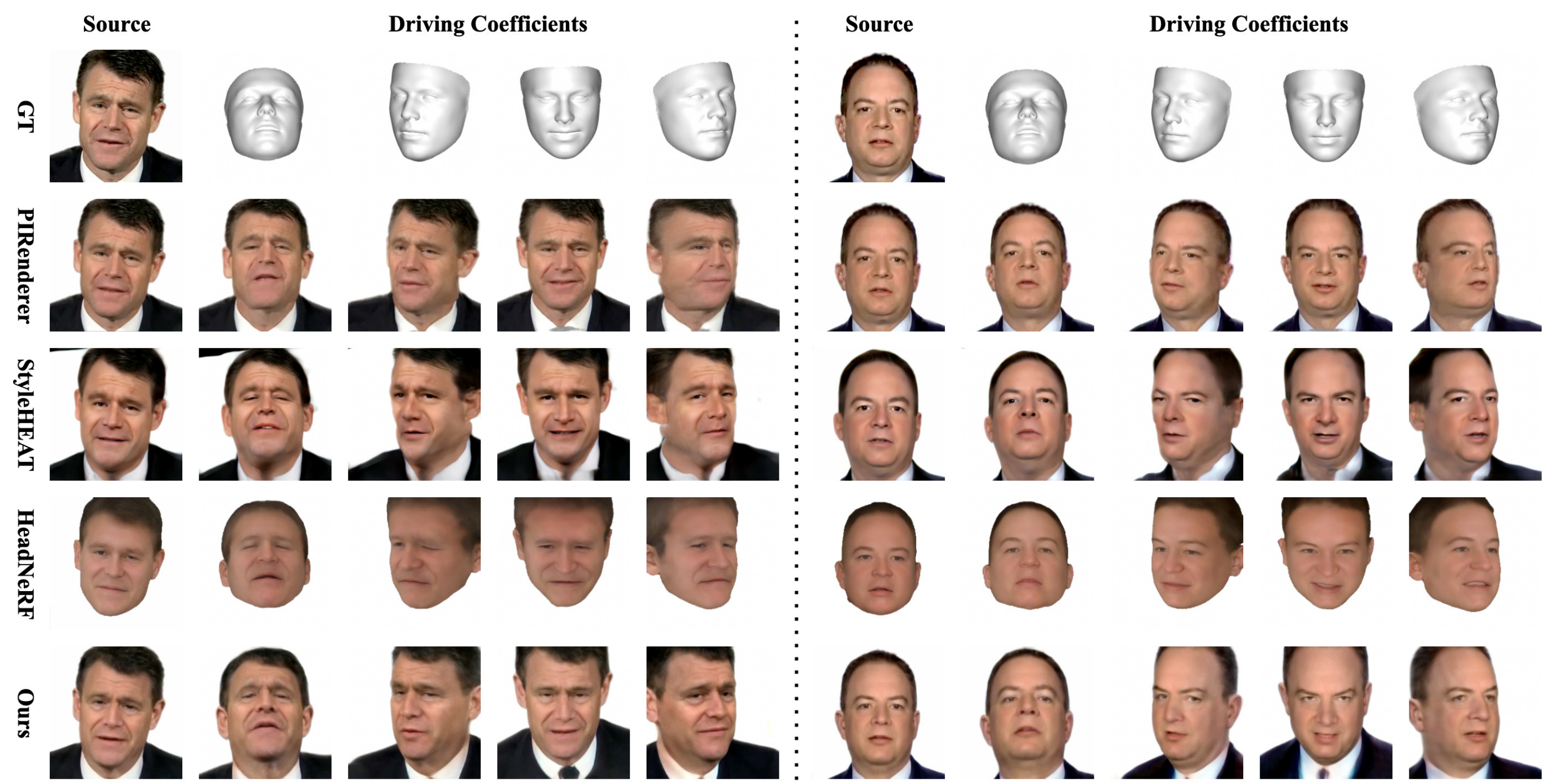}
    \caption{\textbf{3D Consistency on monocular dataset}. We demonstrate additional visualization comparison on the HDTF dataset~\cite{zhang2021flow}, with more drastic camera view variation. All methods use the source frame to extract the identity feature, then use the coefficients of pose and expression as visualized by the face meshes to generate talking faces. This subject is not included in the training set of any methods.}
    \label{fig:hdtf_multiview}
\end{figure*}

\textbf{Monocular dataset}. To evaluate the 3D consistency in the monocular talking dataset, we change the pose coefficients to be rotating while keeping the expression coefficients in sync with the driving frames, as shown in Fig.~\ref{fig:hdtf_multiview}. 

 From both comparisons, we observe PIRenderer~\cite{ren2021pirenderer} and StyleHEAT~\cite{yin2022styleheat} suffer from drastic unnatural image distortion, while our methods can maintain the multi-view consistency and depict natural motion on the expression; compared to the 3D method of HeadNeRF~\cite{hong2022headnerf}, we achieve a more faithful reconstruction of the subject on the skin color and torso. For more detail, please refer to the supplementary video.

\subsection{The Network Architecture of the Controller}

We proceed to describe the controller architecture in this section. Fig.~\ref{fig:controller} shows the details of the motion controller $\emph{C}$. We input the window of adjacent 27 frames of expression and pose coefficients, they form up the motion signal of size 73$\times$27. We use three 1D convolution layers, noted as $\emph{F}_T$ to compress the noises in the sequential 3DMM coefficients. After that, a five-layer MLP is implemented to transform the feature into the magnitudes of 20 orthogonal bases of size 512 in the code book. We calculate 20$\times$14 of such magnitudes, by first outputting 20$\times$15 scalars and then adding 20 of them to the others. Then the 20$\times$14 scalars are multiplied with the orthogonal bases in the code book $\emph{D}$ and transformed to the motion code of size 14$\times$512 . Here 14 is the maximum number of latent codes $\mathbf{w} \in \mathcal{W}$ as input to the face generator $G$~\cite{Chan2022}. The 14$\times$512 latent features form up the $\mathcal{W}^+$ space~\cite{xia2022gan}. In our implementation, it is formulated as the summation of the identity code $\mathbf{w}_{id}$ and the motion code $\mathbf{w}_x$.

    we conduct experiments to evaluate the effectiveness of the proposed controller architecture. One ablation model is constructed by reducing the number of output scalars to 1$\times$20. It is multiplied with the code book $\emph{D}$ to make up the single motion code of size 1$\times$512, it is further repeated 14 times to fit the required shape of 14$\times$512. This model is written as $\mathbf{w}_x \in \mathcal{W}$ in Table.~\ref{tab:ablation_on_controller}. In another ablation model, we replace the code book with vanilla 20$\times$512 linear weights which can also transform every 20 scalars to 512-dimensional latent code. From Table.~\ref{tab:ablation_on_controller}, we observe the performance deterioration on both ablation models on the motion controllability, which is evaluated by AED, APD, and AKD, and also on the image quality as quantified by FID. Even though representing motion code in $\mathcal{W}$ facilitates higher identity consistency as measured by CSIM, it neglects the fact that each of the 14 latent codes contributes differently to the motion deformation, as is testified in 2D GANs~\cite{chong2021stylegan}, therefore the motion controllability is reduced.


\subsection{User Study}

\textcolor{black}{To assess the quality of the animation, we conduct a user study. We sent the results animated by our method and baselines to 20 people (the students in the university). Users are asked to evaluate the animation based on 1) motion that is in sync with driving videos and, and 2) identity similarity compared with the source portrait. Their ratings are averaged, scaled to a maximum of 10, and shown in Table. \ref{tab:user study}. Our method is the most desired in motion controllability and identity consistency.}

\begin{table}
\centering
    \setlength{\tabcolsep}{0.5cm}
        \begin{tabular}{ccc}
        \toprule
              &        Motion $\uparrow$   & Identity $\uparrow$  \\ 
        \midrule
        FOMM\cite{siarohin2019first}   &     4.1 & 5.3 \\
        PIRender\cite{ren2021pirenderer} &     3.5 & 4.3 \\
        StyleHEAT\cite{yin2022styleheat}      & 7.4 & 5.7 \\
        HeadNeRF\cite{hong2022headnerf}      & 4.6 & 3.8 \\
        Ours & \textbf{9.0} & \textbf{8.5} \\ 
        \bottomrule
        \end{tabular}        
        \caption{\textbf{User study on the animation quaility}. 
        }
        \label{tab:user study}
\end{table}

\end{document}